\documentclass{article}

\usepackage[preprint]{neurips_2022}

\usepackage[utf8]{inputenc} 
\usepackage[T1]{fontenc}    
\usepackage{hyperref}       
\usepackage{url}            
\usepackage{booktabs}       
\usepackage{amsfonts}       
\usepackage{nicefrac}       
\usepackage{microtype}      
\usepackage{xcolor}         
\usepackage{graphicx}
\usepackage{amsmath}
\usepackage{amssymb}
\usepackage{booktabs}
\usepackage{epsfig}
\usepackage{graphicx}
\usepackage{amsmath}
\usepackage{amssymb}
\usepackage{amsfonts}
\usepackage{adjustbox}
\usepackage{multicol}
\usepackage{multirow}
\usepackage{booktabs}
\usepackage{rotating}
\usepackage{color}
\usepackage{bbold}
\usepackage{bm}

\title{Multi-View Active Fine-Grained Recognition}

\author{
  Ruoyi Du$^{\ddagger}$, Wenqing Yu$^{\ddagger}$, Heqing Wang$^{\ddagger}$, Dongliang Chang$^{\ddagger}$, \\ \textbf{Ting-En Lin, Yongbin Li, and Zhanyu Ma$^{\ddagger}$}\\
  $^{\ddagger}$Pattern Recognition and Intelligent System Laboratory, School of Artificial Intelligence\\
  Beijing University of Posts and Telecommunications, Beijing $100876$ \\
  \texttt{\{duruoyi, yuwenqing, wangheqing, changdongliang, mazhanyu\}@bupt.edu.cn} \\
}

\begin{document}
\maketitle
\vspace{-0.3cm}
\begin{abstract}
  As fine-grained visual classification (FGVC) being developed for decades, great works related have exposed a key direction -- finding discriminative local regions and revealing subtle differences. However, unlike identifying visual contents within static images, for recognizing objects in the real physical world, discriminative information is not only present within seen local regions but also hides in other unseen perspectives. In other words, in addition to focusing on the distinguishable part from the whole, for efficient and accurate recognition, it is required to infer the key perspective with a few glances, \emph{e.g.}, people may recognize a ``Benz AMG GT'' with a glance of its front and then know that taking a look at its exhaust pipe can help to tell which year's model it is. In this paper, back to reality, we put forward the problem of active fine-grained recognition (AFGR) and complete this study in three steps: (i) a hierarchical, multi-view, fine-grained vehicle dataset is collected as the testbed, (ii) a simple experiment is designed to verify that different perspectives contribute differently for FGVC and different categories own different discriminative perspective, (iii) a policy-gradient-based framework is adopted to achieve efficient recognition with active view selection. Comprehensive experiments demonstrate that the proposed method delivers a better performance-efficient trade-off than previous FGVC methods and advanced neural networks. Codes are available at: \url{https://github.com/PRIS-CV/AFGR}.
\end{abstract}

\section{Introduction}\label{intro}

Aiming at recognizing the sub-categories of objects belong to the same class, in the past two decades, research on fine-grained visual classification (FGVC) has yielded extensive outstanding arts~\cite{lin2015bilinear,xiao2015application,fu2017look,wang2018learning,chen2019destruction,du2020fine,chang2021your,du2021progressive} that surpass human experts in many application scenarios, \emph{e.g.,} recognizing cars~\cite{krause20133d,yang2015large}, aircraft~\cite{maji2013fine}, birds~\cite{wah2011caltech,van2018inaturalist}, and foods~\cite{min2021large}. Despite the great success, the previous efforts on FGVC largely remain limited to a single-view-based paradigm, \emph{i.e.}, identifying the visual content within one single static image. This paradigm may be sufficient for coarse-grained classification where the saturated inter-class differences are easy to capture (\emph{e.g.}, one can distinguish a coupe from other vehicles by its streamlined body, seductive engine, or headlamps). However, things are different for the fine-grained classification scenario where discriminative clues are rare -- one can only dig the subtle structural differences of exhaust pipes to distinguish between different years' models of ``Benz AMG GT'', and there is no other way. Predictably, for single-view-based approaches, an image (view) without discriminative clues existing is completely indistinguishable at the fine-grained level, which fundamentally limits the model's theoretical performance.

\begin{figure*}[t]
\centering
\includegraphics[width=1\linewidth]{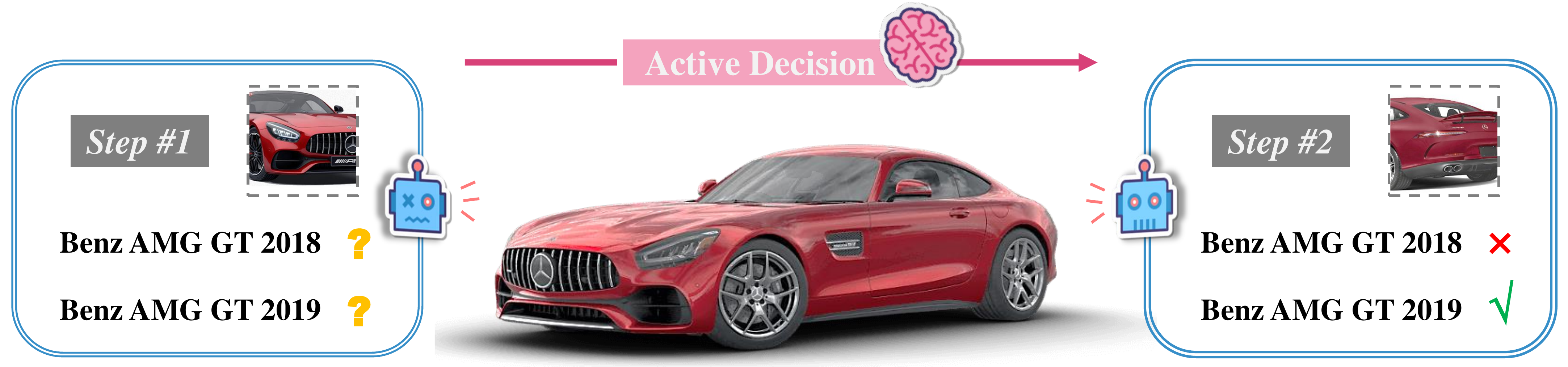}
\caption{Process of the proposed active fine-grained recognition (AFGR). Instead of visiting all possible views, the model is able to infer the next discriminative view according to existing visual information. Here we take two steps for a brief illustration.}
\label{fig:illustration}
\vspace{-0.5cm}
\end{figure*}

Factually, visual recognition is never limited to observing $2$D environments and processing static images. Vision algorithms equipped by portal devices (\emph{e.g.}, smartphone, smart glasses, etc.) or embodied AI agents~\cite{franklin1997autonomous} (\emph{e.g.}, intelligent robots) play the core roles during machine-environment interaction and have become one of the focuses of computer vision research. Therefore, to embrace the new trend, a natural extension of ordinary FGVC 
follows -- in addition to locating discriminative parts within an image, we aim to infer the unseen distinguishable perspective within the physical world ($3$D environment). As shown in Figure~\ref{fig:illustration}, with a single glance from the front, the algorithm may be confused about which year's model the ``Benz AMG GT'' is but can infer that looking at its back will help. 

Specifically, we re-propose the concept of active vision~\cite{aloimonos1988active} in the context of FGVC termed active fine-grained recognition (AFGR) with two essential hypotheses. Firstly, the discriminative information hides in various object views for different fine-grained categories, which determines that discriminative perspective inference is non-trivial and worth studying. Secondly, indistinguishable views also contain visual clues leading to the discriminative perspective, which ensures the solvability of the problem.

To start with, due to the absence of qualified datasets, we first collect a hierarchical, fine-grained, multi-view vehicle dataset named Multi-view Cars (MvCars) as the testbed. MvCars contains $20$ models of cars from $4$ brands and covers more than one car type (\emph{e.g.}, coupe, SUV, etc.) for each brand. Furthermore, to ensure the difficulty of MvCars, we include each brand with two similar categories, \emph{e.g.}, different years' models of the same series. There are $7$ aligned views for each car and about five thousand images included in the dataset. Right after that, our first hypothesis is verified(see Section~\ref{dataset}), which indicates that MvCars is sound for the problem raised.

Secondly, our next contribution following is an efficient multi-view fine-grained recognition framework via active next-view selection. In particular, following the general idea of view-based $3$D object understanding~\cite{su2015multi}, an extraction-aggregation architecture is designed as the feature encoder, where a convolutional neural network (\emph{e.g.}, ResNet$50$~\cite{he2016deep}) is first applied to extract single-view features independently, and then a recurrent neural network (\emph{e.g.}, GRU~\cite{chung2014empirical}) is adopted to aggregate multi-view features and form global descriptions. Afterward, we formulate the next-view selection as a sequential decision process, where the model is demanded to {\it decide the next discriminative view} (action) according to {\it previously observed views} (state). Thus, a  proximal policy optimization (PPO)~\cite{schulman2017proximal} is implemented and revised for training. Note that the proposed framework does not rely on specific neural network architectures. It can extend any visual recognition network to an FGVC expert in the 3D environment.

Finally, several carefully designed baselines are re-produced on MvCars as benchmark results, including general neural networks in a multi-view recognition setting, and popular FGVC methods. Instead of time costs/computation budgets, we adopt the required step numbers for reliable prediction to measure the model efficiency. This is because the time cost for acquiring one more view far outweighs the inference cost, and it may need users' efforts (for applications on portal devices). The experimental results demonstrate that the proposed method delivers a better performance-efficient trade-off than all competitors. After that, an analysis of the upper bound of the proposed method reveals the FGVC characteristic inherited by AFGR. In addition, comprehensive ablation studies are carried out to verify the necessities of each model component.




\section{Related Work}

\subsection{Fine-Grained Visual Classification}

Due to the inherent subtle inter-class variance and the relatively large intra-class variance, fine-grained visual classification is much more challenging than ordinary coarse-grained classification. With vigorous efforts made by researchers, great progress has been made in many directions. Localization based approaches~\cite{zhang2014part,krause2015fine,xiao2015application,fu2017look,wang2018learning} that explicitly locate discriminative parts for feature extraction to alleviate the intra-class variance. High-order encoding methods~\cite{lin2015bilinear,gao2016compact,fu2017look,yu2018hierarchical,zheng2019learning} that adopt high-order feature interactions for better representation ability that can capture subtle difference. Chen \emph{et al.}~\cite{chen2019destruction} and Du \emph{et al.}~\cite{du2020fine} train the model with jigsaw patches to implicitly encourage knowledge mining from local regions. Recently, Chang \emph{et al.}~\cite{chang2021your} leverage the underlying hierarchical structure of fine-grained categories to achieve user-friendly outputs and better performance.

Except for good performances being brought, these aforementioned works also reveal that FGVC is never just a harder classification problem but a stand-alone field that requires well-directed research. In this paper, to further broaden the horizon of FGVC, we propose the active fine-grained recognition (AFGR) task aiming at effective recognition of fine-grained categories in the $3$D environment along with a targeted dataset. It is worth noting that the CompCars~\cite{yang2015large} dataset also provides a car dataset with view annotations. However, its multi-view images are taken from different objects, making it less suitable for the raised problem. 

\vspace{-0.2cm}
\subsection{Multi-View Recognition}

Elsewhere for ordinary object recognition problems, certain progress has been made to recognizing $3$D objects with three streams can be summarized~\cite{chen2021mvt}: point-based methods~\cite{qi2017pointnet,qi2017pointnet++,aoki2019pointnetlk,ni2020pointnet++}, volume-based methods~\cite{maturana2015voxnet,wu20153d,qi2016volumetric,meng2019vv}, and view-based methods~\cite{chiu2007virtual,su2015multi,johns2016pairwise,kanezaki2018rotationnet,yu2018multi,yang2019learning}. Among them, point-based and volume-based approaches demand to perceive the $3$D structure of objects via lidar, depth sensor or something else, which makes them less practicable in daliy applications, \emph{e.g.}, recognizing an unfamiliar car for detailed information simply with a mobile phone. On the contrary, view-based methods that leverage multiple surrounding $2$D views as descriptors for $3$D objects tend to be an optimal choice. 

Specifically, view-based methods share the core idea that encoding single-view features through vision neural network and then aggregating multi-view features. Su \emph{et al.}~\cite{su2015multi} first approaches the multi-view recognition problem with CNN for feature extraction and sum-pooling for aggregation. Then, Johns \emph{et al.}~\cite{johns2016pairwise} decomposes image sequences into image pair sets, and then aggregates the pair-based classification in a weighted manner. After that, feature concatenation~\cite{wang2019dominant}, hierarchical attention~\cite{han20193d2seqviews}, and weighted fusion~\cite{feng2018gvcnn} are also adopted for better aggregating sequence features. In addition, sequences models (\emph{e.g.}, LSTM~\cite{hochreiter1997long}, GRU~\cite{chung2014empirical}, Transformer blocks~\cite{vaswani2017attention}, etc) are also widely considered~\cite{jayaraman2016look,han2018seqviews2seqlabels,chen2021mvt} and demonstrate their effectiveness. 

In this paper, specifically towards the active fine-grained recognition (AFGR) task we raised, traditional multi-view recognition dataset (\emph{e.g.}, RGB-D~\cite{lai2011large}, ModelNet10, ModelNet40~\cite{wu20153d}) is not sufficient any more. Thus, we first collect a fine-grained, multi-view vehicle dataset named MvCars as our testbed. Then, an active fine-grained recognition framework is built upon the general extraction-aggregation scheme. Note that, similar to ours, some approaches also take recognition efficiency into consideration~\cite{jayaraman2016look,johns2016pairwise} by actively controlling the agent motion within a viewing sphere. While a strict viewing sphere is not readily available in daily applications, especially for recognition with portable devices, hence we consider the view selection as a discontinuous classification problem here.


\section{Methodology}

\begin{figure*}[t]
\centering
\includegraphics[width=1\linewidth]{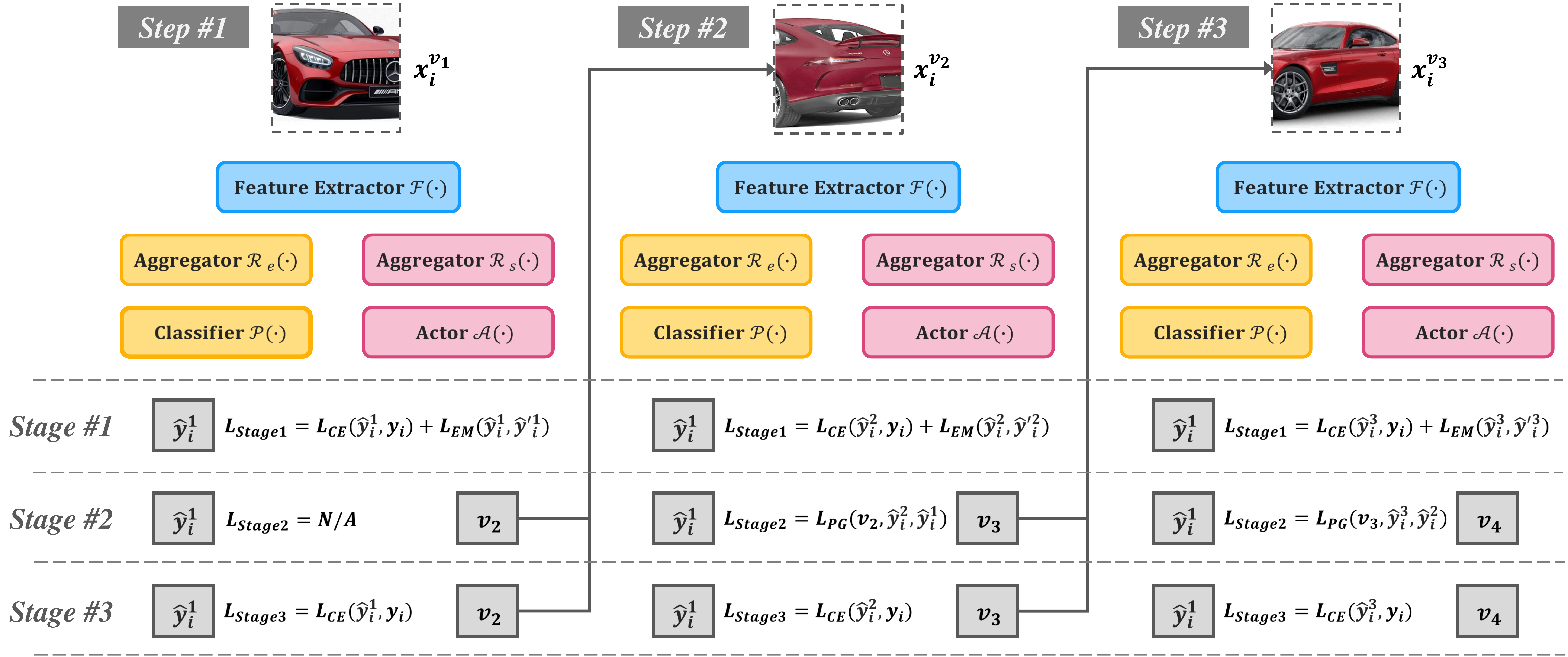}
\caption{Illustration of the proposed AFGR framework where three training stages are included: \textbf{Stage I} for training a multi-view recognition model with smooth predictions, \textbf{Stage II} for optimizing the next-view selection component based on the behavior of the classifier, and \textbf{Stage III} for fine-tuning the recognition model along with the trajectory decided by the actor. Here we use three training steps for brief illustration.}
\label{fig:framework}
\vspace{-0.3cm}
\end{figure*}

\subsection{Overview}

Here we first give an overview of data flow during inference along with the setting of active fine-grained recognition (AFGR). 

\textbf{Data structure.} For AFGR, a dataset consists of $N$ samples can be expressed as $\{X_i,y_i\}_{i=1}^N$, where $X_i=\{x_i^1,\dots,x_i^v,\dots,x_i^V\}$ is a sequence of images depict a specific sample from $V$ perspectives and $y_i$ is their common ground-truth label. Note that, for arbitrary two samples $X_i$ and $X_j$, $x_i^v$ and $x_j^v$ are taken from the same perspective, which means the annotations of views $\{1,\dots,V\}$ are aligned.

\textbf{Inference process.} For sample $X_i$, the model will take an image $x_i^{v_1}$ from arbitrary view $v_1$ as the initial visual input, which simulates the situation that the model may start recognition while facing any views of the target object. After that, the recognition process will carry on step-by-step. In particular, at step $t$ with input $x_i^{v_t}$ from view $v_t$, the model will utilize all currently perceived information $\{x_i^{v_1},\dots,x_i^{v_{t-1}},x_i^{v_t}\}$ to deliver the category prediction $\hat{y}_i^t$ and the next-view proposal $v_{t+1}$. Then, a inference cycle is closed, and the process can keep going with $x_i^{v_{t+1}}$ as the next input. 

\textbf{Framework component.} To process a sequence of correlated visual inputs, an extraction-aggregation structure tends to be an intuitive choice. Specifically, for any image $x_i^{v_t}$ input the system, a CNN-based feature extractor $\mathcal{F}(\cdot)$ is first applied to extract single-view feature as $f_i^{v_t}=\mathcal{F}(x_i^{v_t})$. It is worth reminding that the feature extractors for different views share their weights and this design will not leads to additional parameters. After that, an ideal model should take all previously acquired information into consider. Thus, a recurrent neural network is introduced as the aggregator $\mathcal{R}(\cdot)$ that aggregates features from all seen perspectives. In particular, here we adopts two aggregator with same structure but individual weights $\mathcal{R}_e(\cdot)$ and $\mathcal{R}_s(\cdot)$ that, $\mathcal{R}_e(\cdot)$ form global embeddings $e_i^t=\mathcal{R}_e(f_i^{v_1},\dots,f_i^{v_t})$ for category prediction, while $\mathcal{R}_s(\cdot)$ depicts the current states $s_i^t=\mathcal{R}_s(f_i^{v_1},\dots,f_i^{v_t})$ for next-view selection. Finally, a classifier $\mathcal{P}(\cdot)$ and an actor $\mathcal{A}(\cdot)$ are equipped in parallel with outputs $\hat{y}_i^t=\mathcal{P}(e_i^t)$ and $v_t=\mathcal{A}(s_i^t)$, respectively.

\subsection{Model Training}

According to the aforementioned inference process, we can tell that the recognition component and the next-view selection component work in a separate but not independent manner. The mission of the recognition component is quite straightforward -- conducting category prediction based on acquired information as well as possible. While the optimization goal of the next-view selection component largely depends on the behavior of recognition -- basically, the actor should try to select the next-view that can maximize the prediction probability of the target category. Therefore, a three stages training framework is intuitively designed: \textbf{Stage I} aims to train a good recognition model (including $\mathcal{F}(\cdot)$, $\mathcal{R}_e(\cdot)$, and $\mathcal{P}(\cdot)$) that can handle sequence input, \textbf{Stage II} aims to optimize next-view selection (where $\mathcal{R}_s(\cdot)$ and $\mathcal{A}(\cdot)$ participate) according to the behavior of the trained recognition model, and \textbf{Stages III} aims to refine the recognition model under the trajectories decided by the actor. The whole framework is illustrated in Figure~\ref{fig:framework}, and introductions about the three stages are as follows.

\textbf{Stage I.} We first train a recognition model that can handle a sequence of inputs with dynamic length. Each training iteration is divided into $T$ steps with input sequence lengths from $1$ to $T$. For the $t$-th step, a new image $x_i^{v_t}$ is randomly selected from unseen views and appended to the input sequence at the $(t-1)$-th step. Here we set $T=V$ to ensure the sequence is no-duplicated. Thus, with cross-entropy for optimization, the loss function for a batch of $B$ samples can be formulated as:
\begin{equation}
L_{CE}(\hat{y}_i^t,y_i)=\frac{-1}{BT}\sum_{i=1}^B\sum_{t=1}^T y_i \times log(\hat{y}_i^t).
\end{equation}
Note that the inductive bias behind training the recognition component in the first place is that its behavior can reveal view discrimination -- a more discriminative view will greatly reduce the entropy of category prediction. However, a well-convergent classification model often tends to deliver high confidence predictions, especially for the small-scale datasets in the FGVC scenario, which will cause little changes in prediction probabilities and limit the information being revealed. Therefore, we further introduce an entropy maximization constraint to encourage smooth predictions. Specifically, let $p_i^t$ be the output of the classifier before the softmax function. A softer version of the prediction can be obtained by introducing a pre-defined temperature $h$, which is expressed as: 
\begin{equation}
\hat{y}_{i,j}^{'t}=\frac{exp(p_{i,j}^t)/h}{\sum_k exp(p_{i,k}^t)/h},
\end{equation}
where $j$ and $k$ indicate channel index of $p_i^t$. Then, we minimize the Euclidean Distance between $\hat{y}_i^t$ and $\hat{y}_i^{'t}$ to achieve entropy maximization as:
\begin{equation}
L_{EM}(\hat{y}_i^t,\hat{y}_i^{'t})=\frac{-1}{BT}\sum_{i=1}^B\sum_{t=1}^T |\hat{y}_i^t - \hat{y}_i^{'t}|^2.
\end{equation}
The total loss of \textbf{Stgae I} is $L_{Stage1}=L_{CE}(\hat{y}_i^t,y_i)+L_{EM}(\hat{y}_i^t,\hat{y}_i^{'t})$, and the degree of entropy maximization constrain can be control by different temperature $h$.

\textbf{Stage II.} Here, the recognition components ($\mathcal{F}(\cdot)$, $\mathcal{R}_e(\cdot)$, and $\mathcal{P}(\cdot)$) are frozen, and we only optimize $\mathcal{R}_s(\cdot)$ and $\mathcal{A}(\cdot)$ for next-view selection. As a sequential decision problem, we adopt policy gradient method for optimization instead of directly optimizing with the classification loss, since the view selection process is non-differentiable. At the $t$-th ($t\ge2$) training step, the model will receive the input $x_i^{v_t}$ with the perspective $v_t$ decide by the actor at the $t-1$-th step. Then the view selection components can be updated according to the change of target category prediction probability, \emph{i.e.}, the rewards is set as $r_i^t=\hat{y}_{iy_i}^t-\hat{y}_{iy_i}^{t-1}$. And the $t$-th ($t\ge2$) step's loss function of \textbf{Stage II} can be simply expressed as: $L_{stage2}=L_{PG}(v_t,\hat{y}_{i}^{t-1},\hat{y}_{i}^{t})$.

It is worth noting that, for popular policy gradient algorithms~\cite{schulman2016high,schulman2017proximal}, the total reward for the current step's optimization is a (weighted) sum of all feature rewards from now on. This is because these methods are designed for scenarios where an agent is required to achieve an ultimate goal through a series of actions. However, on the contrary, AFGR aims at using as few steps as possible to achieve as high accuracy as possible, \emph{i.e.}, we care more about how to achieve the best performance at the current step rather than in the future. Therefore, we slightly modify the policy gradient algorithm by utilizing only $r_i^t$ for the $t$-th step's optimization. 

\textbf{Stage III.} There is nothing new in this stage, all settings are the same as \textbf{Stage I} except for (i) the selected view $v_t$ when $t\ge2$ is given by the actor, and (ii) the entropy maximization constraint is removed (\emph{i.e.}, $L_{Stage1}=L_{CE}(\hat{y}_i^t,y_i)$). We hope the model can be refined under standard classification supervision (\emph{i.e.}, purely with the cross-entropy loss) to especially adjust the trajectories decided by the actor. 

\subsection{Design Details}

\textbf{Feature extractor $\mathcal{F}(\cdot)$.} The feature extractor can be any backbone network for vision tasks, including various CNN architectures and Transformers. Besides, by replacing $\mathcal{F}(\cdot)$ with other FGVC models, the proposed method can also extend them to work in $3$D environments.

\textbf{Feature aggregator $\mathcal{R}_e(\cdot)$ and $\mathcal{R}_s(\cdot)$.} The two feature aggregators should be able to aggregate information from sequences with variable lengths. Here we adopt GRU~\cite{chung2014empirical} for best performance. There are also alternatives like LSTM~\cite{hochreiter1997long}, self-attention block~\cite{vaswani2017attention}, etc, which we will discuss in Appendix~B.

\textbf{Classifier $\mathcal{P}(\cdot)$ and actor $\mathcal{A}(\cdot)$.} Both the classifier and the actor are formed by one fully connected layer. For the cases that equip the proposed framework with other FGVC approaches, the structure of the classifier can be modified accordingly.

\textbf{Policy gradient algorithm.} We adopt the proximal policy optimization (PPO)~\cite{schulman2017proximal} for the training of next-view selection with the reward of the current step only. Details can be found in Appendix~A.

\section{Dataset}\label{dataset}

\begin{figure*}[htbp]
\centering
\includegraphics[width=1\linewidth]{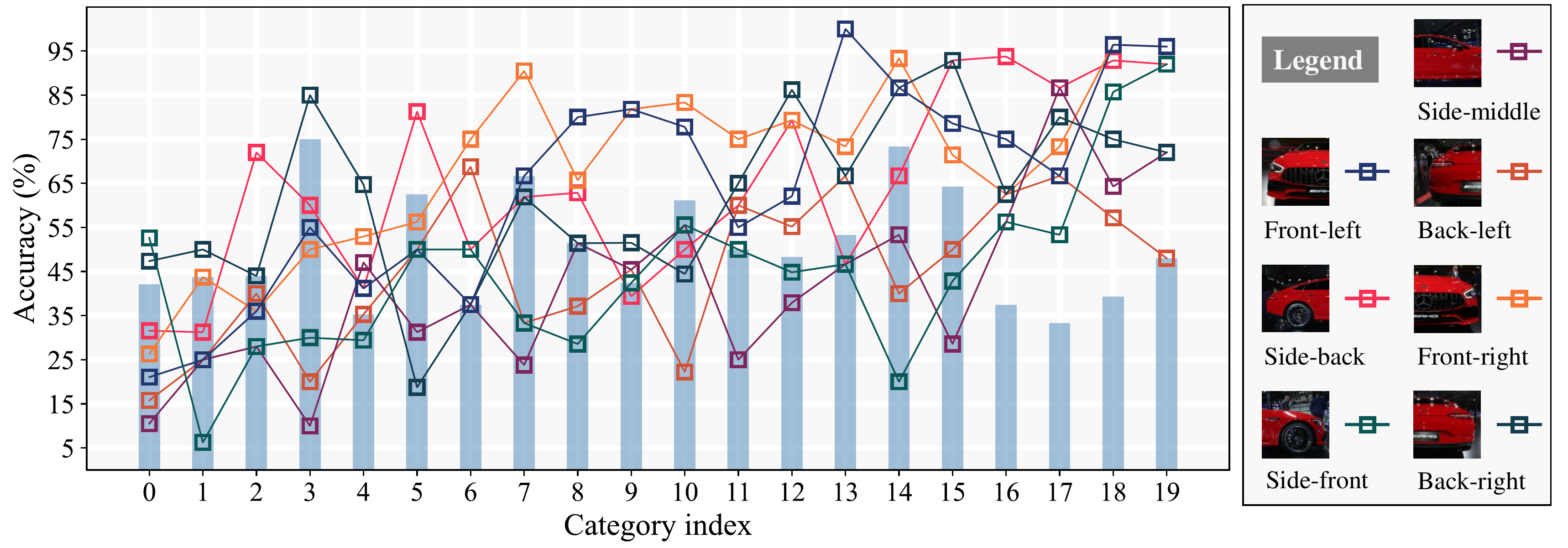}
\caption{The quantitative analysis of the collected MvCars dataset. The broken-lines show model accuracy based on $7$ individual views. And the bars represent the differences between the maximum and minimum accuracy for each category.}
\label{fig:view_vs_acc}
\vspace{-0.2cm}
\end{figure*}

\textbf{Data collection and statistic.} The Multi-view Cars (MvCars) dataset is collected from $4$ automobile sale sites\footnote{1.\url{www.autohome.com}\\ \indent\indent2.\url{www.yiche.com}\\ \indent\indent3.\url{www.dongchedi.com}\\ \indent\indent4.\url{www.pcauto.com.cn}} where cars are displayed from different perspectives. To ensure the diversity and representativeness of MvCars, we choose $20$ models of cars from $4$ popular brands (Mercedes-Benz, Volkswagen, Toyota, and Nissan) where each brand contains cars of at least $2$ types (\emph{e.g.}, coupe, SUV, etc). For each car, we annotated $7$ aligned perspectives -- front-left, front-right, side-front, side-middle, side-back, back-left, and back-right, and samples with missing perspectives are discarded. In total, there are $4669$ images collected and then split into $2450$/$2219$ for train/test set, respectively.

\textbf{Quality verification.} With the collected MvCars, here we first experimentally validate our first hypothesis mentioned in Section~\ref{intro} -- the discriminative information hides in various object views for different fine-grained categories. Factually, it is two-fold: (i) different perspectives contribute differently to FGVC, otherwise, actively selecting object view is meaningless, and (ii) different categories own different discriminative perspectives, otherwise, there is a trivial solution existing -- consistently seeking the fixed distinguishable view.

In particular, for each perspective, we train a ResNet50~\cite{he2016deep} for classification and obtain its accuracy in each category. Therefore, for any specific category, we can tell which perspective is more distinguishable by comparing the performances of $7$ models based on different views. The experimental results are shown in Figure~\ref{fig:view_vs_acc}. Bars in the graph indicate the differences between the maximum and minimum accuracy of each category, where we can observe that the differences are about $50.5\%$ on average and at least more than $30.3\%$. It powerfully proves that different perspectives contribute differently in the context of FGVC. On the other hand, broken-lines in the graph represent view accuracy changes along with different categories. The interaction of lines indicates that the ranking of view discrimination is not consistent, demonstrating that different categories have different discriminative perspectives. 

In one word, in MvCars, different perspectives provide significantly various meanings for FGVC, which is also hard to pre-defined via prior knowledge. Thus, an active recognition method is called for, and the collected MvCars dataset can serve as an eligible testbed.

\vspace{-0.1cm}
\section{Experiment}
\vspace{-0.1cm}

\begin{table*}[t]
\caption{Results of the proposed method against different baselines. The table is divided into five sections according to different backbones, and the best results of each section are marked in \textbf{bold}.}
\vspace{0.3cm}
\label{tbl:main}
\begin{adjustbox}{width=0.95\linewidth,center}
\begin{tabular}{c c c c c}
\toprule[1pt]
\textbf{Method} & \textbf{Backbone} & mAcc. ($\%$) & w-mAcc. ($\%$) & Step$2$-Acc. ($\%$)\\
\midrule[1pt]
Hierarchical BCNN & ResNet$50$ & $81.37\pm0.2$ & $81.57\pm0.3$ & $80.90\pm0.6$\\
Pairwise Confusion & ResNet$50$ & $82.35\pm0.5$ & $82.25\pm0.6$ & $81.48\pm0.5$\\
CrossX & ResNet$50$ & $84.92\pm0.4$ & $85.13\pm0.3$ & $84.79\pm0.3$\\
PMG & ResNet$50$ & $84.68\pm0.5$ & $84.72\pm0.6$ & $84.10\pm0.6$\\
CAL & ResNet$50$ & $84.23\pm0.1$ & $84.54\pm0.3$ & $84.39\pm0.5$\\
Sequence Baseline & ResNet$50$ & $84.18\pm1.2$ & $84.56\pm1.3$ & $83.05\pm1.7$\\
Ours & ResNet$50$ & $\mathbf{86.07\pm0.5}$ & $\mathbf{87.66\pm0.5}$ & $\mathbf{87.20\pm0.4}$\\
\midrule[0.5pt]
Sequence Baseline & DenseNet$169$ & $85.43\pm0.7$ & $85.70\pm0.7$ & $85.07\pm1.0$\\
Ours & DenseNet$169$ & $\mathbf{85.92\pm0.9}$ & $\mathbf{86.53\pm0.8}$ & $\mathbf{86.03\pm0.6}$\\
\midrule[0.5pt]
Sequence Baseline & EfficientNet\_b$3$ & $83.70\pm0.6$ & $83.85\pm0.5$ & $81.98\pm0.4$\\
Ours & EfficientNet\_b$3$ & $\mathbf{84.67\pm0.3}$ & $\mathbf{85.27\pm0.6}$ & $\mathbf{83.91\pm0.6}$\\
\midrule[0.5pt]
Sequence Baseline & RegNetY\_$1.6$GF & $84.38\pm0.1$ & $84.75\pm0.5$ & $83.91\pm1.0$\\
Ours & RegNetY\_$1.6$GF & $\mathbf{84.75\pm0.1}$ & $\mathbf{85.22\pm0.1}$ & $\mathbf{84.44\pm0.2}$\\
\midrule[0.5pt]
TransFG & ViT-B\_$16$ & $81.18\pm0.4$ & $81.09\pm0.4$ & $80.33\pm0.2$\\
Sequence Baseline & ViT-B\_$16$ & $80.61\pm0.7$ & $81.26\pm0.9$ & $79.96\pm0.9$\\
Ours & ViT-B\_$16$ & $\mathbf{81.44\pm0.8}$ & $\mathbf{82.23\pm1.0}$ & $\mathbf{81.30\pm0.9}$\\
\midrule[1pt]
\end{tabular}
\end{adjustbox}
\vspace{-0.3cm}
\end{table*}

In this section, first, we introduce the baseline models for comparison and the metrics for evaluation. Then we discuss the comparison results in Section~\ref{main_results}. After that, we discuss the performance upper bound of our model in Section~\ref{upper_bound_analysis}. Finally, ablation studies are carried out in Section~\ref{ablation_studies} to verify our design choices. In addition, the implementation details can be found in Appendix~A., and additional ablation studies about hyper-parameters and network architectures can be found in Appendix~B. 

\textbf{Baseline models.} For extensively evaluation, two groups of baseline methods are designed and implemented. The first is state-of-the-art FGVC methods, including Hierarchical BCNN~\cite{yu2018hierarchical}, Pairwise Confusion~\cite{dubey2018pairwise}, CrossX~\cite{luo2019cross}, PMG~\cite{du2020fine}, CAL~\cite{rao2021counterfactual}, and TransFG~\cite{he2021transfg}. To extend these approaches to the multi-view recognition scenario, we employ a naive model ensemble scheme, \emph{i.e.}, at the $t$-th step, the average of $t$ inputs' predictions is adopted as the current result. The second group is advanced vision neural networks, including ResNet~\cite{he2016deep}, DenseNet~\cite{huang2017densely}, EfficientNet~\cite{tan2019efficientnet}, RegNet-Y~\cite{radosavovic2020designing}, and ViT~\cite{dosovitskiy2020image}. Due to their conciseness (no complicated training strategies or carefully designed structures), we can easily implement them in the extraction-aggregation form (more general for multi-view recognition~\cite{jayaraman2016look,han2018seqviews2seqlabels,chen2021mvt}) with GRU~\cite{chung2014empirical} for feature aggregation. The second sequence-based baseline group is also used to demonstrate the generalization ability of the proposed framework by serving as the recognition model trained in \textbf{Stage I}. Note that, for these baseline methods, the input of each step is randomly selected with no duplicate view.

\textbf{Evaluation Metrics.} For quantitative evaluation, results based on $3$ metrics are reported: (i) \textbf{Mean Accuracy (mAcc)} that takes the mean value of all $T$ steps' accuracy, which can be regarded as the area under the accuracy-step line that represents the general performances of models, (ii) \textbf{Weighted Mean Accuracy (w-mAcc)} that weights different steps with exponentially decreased weights, since the performance of the first few steps should be more important in the consideration of efficiency\footnote{Here we take $[0.0000, 0.5079, 0.2540, 0.1270, 0.0635, 0.0317, 0.0159]$ for w-mAcc when $T=7$. The accuracy of the first step is weighted by $0.0$ because it is randomly selected and does not relate to the performance of active selection.}, and (iii) \textbf{Step$2$ Accuracy (Step$2$-Acc)} that takes the $2$-nd step's accuracy to highlight the profit of the first view selection\footnote{Step$2$-Acc can be regarded as w-mAcc with weight set $[0.0, 1.0, 0.0, 0.0, 0.0, 0.0, 0.0]$.}. In addition, following~\cite{wang2020glance}, we introduce a dynamic exit strategy to further reveal the model potential under given step expectations -- given the expectation of step number, confidence thresholds for exiting inference at each step are dynamically defined according to the training data, which enables better resource allocation among all test data (details can be found in appendix~A.).


\subsection{Main Results}\label{main_results}

\begin{figure*}[t]
\begin{minipage}[t]{0.47\linewidth}
\centering
\includegraphics[width=1\linewidth]{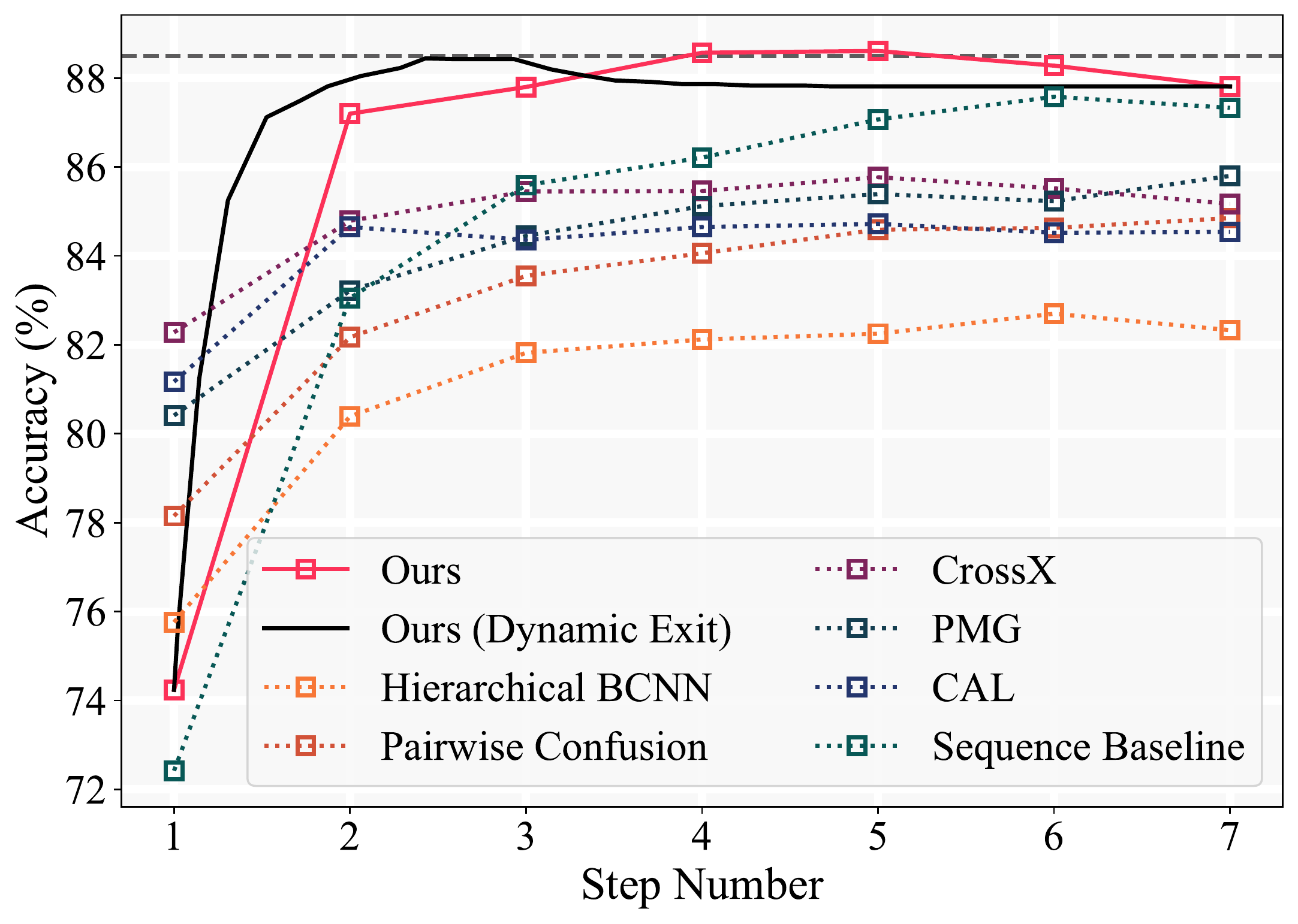}
\caption{Accuracy-step lines/curve of the proposed method against competitors.}
\label{fig:main_exp}
\end{minipage}
\hspace{0.3cm}
\begin{minipage}[t]{0.47\linewidth}
\centering
\includegraphics[width=1\linewidth]{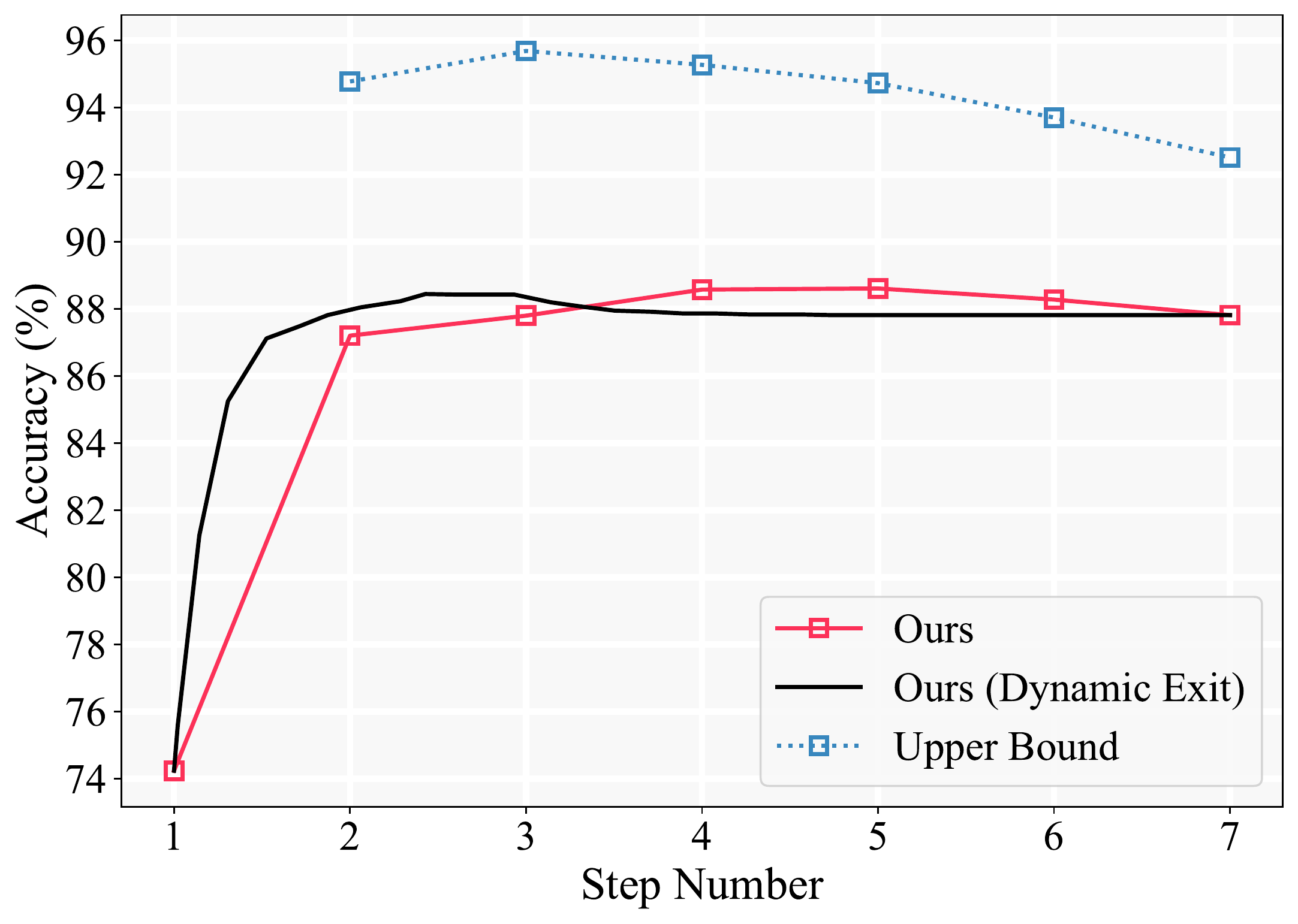}
\caption{The performance upper bound from the perspective of trajectory decision.}
\label{fig:upper_bound}
\end{minipage}
\vspace{-0.3cm}
\end{figure*}

The results of the proposed method against all mentioned baselines are reported in Table~\ref{tbl:main}. The table is organized into $5$ sections according to different backbone networks, and we mainly focus on the comparison within each section for fairness. For ResNet$50$~\cite{he2016deep} as the base model, we can observe that the sequence-based model (sequence baseline in the table), aiming at multi-view recognition, delivers quite competitive results that even consistently surpass FGVC methods like Hierarchical BCNN~\cite{yu2018hierarchical} and Pairwise Confusion~\cite{dubey2018pairwise}. On the contrary, the proposed method outperforms it by $\sim2.\%$, $\sim3.\%$, $\sim4.\%$ for mAcc, w-mAcc, and Step$2$-Acc, respectively. The larger margins on w-mAcc and Step$2$-Acc also demonstrate its superiority in efficiency that benefits from the active next-view selection scheme. Besides, there is no doubt that our framework obtains state-of-the-art performance with any backbone networks, which indicates its robustness and generalization ability.

To better illustrate the change of model accuracy over inference steps, we show the accuracy-step lines of all models with ResNet$50$ as the backbone in Figure~\ref{fig:main_exp}. In addition, we also include the curve formed by the dynamic exit strategy~\cite{wang2020glance} for our model. Firstly, we can observe that when the step number $t=1$, \emph{i.e.}, the prediction is conducted based on a single image, FGVC approaches demonstrate their professionalism by outperforming both the proposed method and the sequence baseline. This is reasonable since the proposed one is just a ResNet$50$-based classification model with random inputs when $t=1$. However, for $t\ge2$, our model immediately dominated the game -- specifically, it surpasses all competitors with significant margins when $t=2$, echoing the results of Step$2$-Acc in Table~\ref{tbl:main}. We attribute this to the effectiveness of our next-view selection mechanism. Last but not least, with a better resource allocation brought by the exit strategy via dynamic sequence length arrangement, a significant further improvement can be observed in the first few steps -- we can obtain the best performance with $\sim2.$ steps less.

At this point, the audiences may question why our model's performance does not consistently increase. With the same question, we study the upper bound of our model in the next subsection.

\subsection{Upper Bound Analysis}\label{upper_bound_analysis}

Due to the finite total view numbers, we are able to visit all possible trajectories for each sample. Therefore, a performance upper bound can be obtained from the perspective of trajectory decision. In particular, given a sequence length, any sample can be regarded as a correctly classified sample long as there exists one trajectory that can yield the correct prediction. As shown in Figure~\ref{fig:upper_bound}, the degradation in the last few steps is also observed on our upper bound. We attribute this to the inherent feature of fine-grained recognition in the $3$D environment -- the discriminative clues only hide in a few views, and the noises caused by intra-class variance will be more likely to be introduced when full visual information (\emph{i.e.}, all views) is included. This particularly echoes the essential insight in the 2$D$ fine-grained recognition where subtle differences of local regions are discriminative, and the global structures are more likely disturbed.

\subsection{Ablation Study}~\label{ablation_studies}
\vspace{-0.5cm}

\begin{table*}
\caption{Results of ablation studies. The best results are marked in \textbf{bold}.}
\vspace{0.3cm}
\label{tbl:abl}
\begin{adjustbox}{width=0.95\linewidth,center}
\begin{tabular}{c c c c c}
\toprule[1pt]
\textbf{Method} & \textbf{Backbone} & mAcc. ($\%$) & w-mAcc. ($\%$) & Step$2$-Acc. ($\%$)\\
\midrule[1pt]
Random Selection & ResNet$50$ & $85.43\pm0.5$ & $86.14\pm0.7$ & $84.87\pm0.8$\\
Allow Duplicate View & ResNet$50$ & $85.33\pm0.3$ & $87.04\pm0.3$ & $86.72\pm0.4$\\
\midrule[0.5pt]
w/o Entropy Maximization & ResNet$50$ & $85.91\pm0.9$ & $87.07\pm0.9$ & $86.07\pm0.9$\\
w/ Future Rewards & ResNet$50$ & $86.01\pm0.6$ & $87.10\pm0.6$ & $86.17\pm0.6$\\
w/o Stage III & ResNet$50$ & $85.31\pm1.6$ & $86.57\pm1.6$ & $85.63\pm1.7$\\
\midrule[0.5pt]
Ours & ResNet$50$ & $\mathbf{86.07\pm0.5}$ & $\mathbf{87.66\pm0.5}$ & $\mathbf{87.20\pm0.4}$\\
\midrule[1pt]
\end{tabular}
\end{adjustbox}
\vspace{-0.4cm}
\end{table*}

In this section, we evaluate several variants of the proposed method based on ResNet$50$ to demonstrate the necessities of our designs. First, to directly verify the effectiveness of the active next-view selection mechanism, we study our model trained via $3$ stages with randomly selected inputs. Fortunately, the proposed method passes the test with significant margins of $\sim0.6\%$, $\sim1.5\%$, and $\sim2.3\%$. Additionally, for all evaluations before, an artificial restriction is added to ensure new views selected are unseen. It is intuitive since unseen views can offer complementary information, and the information about which views have been selected is easily acquired. Here we also evaluate by allowing duplicate views, and the model performance degrades with no surprise. After that, our designs for model training are also demonstrated to be effective. It is worth noting that when we include the future rewards for policy optimization, mAcc is not significantly affected ( with a slight degradation of $0.06\%$), but w-mAcc and Step2-Acc decrease by $\sim0.6\%$ and $\sim1.0\%$. This indicates that future rewards may be meaningful for traditional sequential decision problems but not for AFGR which highly requires efficiency.

\vspace{-0.1cm}
\section{Conclusion}
\vspace{-0.1cm}

In this paper, we extend the fine-grained visual classification to $3$D environments and put forward the active fine-grained recognition (AFGR) problem. A multi-view car dataset (MvCars) is collected as a qualified benchmark. We re-implement several FGVC approaches and several vision neural networks under a general multi-view recognition scheme as baseline methods. A policy-gradient-based framework is introduced for the problem raised. The proposed method yields the best performance on MvCars. We also discuss the upper bound of our framework from the perspective of trajectory decision.

\begin{ack}

\end{ack}


{
\small
\bibliographystyle{plain}
\bibliography{ref.bib}
}

\appendix

\section*{Appendix}

\section{Implementation Details}

\subsection{Implementation of the Policy Gradient Algorithm}

For training the actor $\mathcal{A}(\cdot)$ (the next-view selection module) at \textbf{Stgae II}, we adopt the proximal policy optimization (PPO) algorithm~\cite{schulman2017proximal} with a slight modification. Specifically, given a series of inputs ${x_i^{v_1},\dots,x_i^{v_t}}$ at the $t$-th step, the extractor $\mathcal{F}(\cdot)$ and the aggregator $\mathcal{R}_s(\cdot)$ are first applied to form the current state: 
\begin{equation}
    s_i^t=\mathcal{R}_s(\mathcal{F}(x_i^{v_1}),\dots,\mathcal{F}(x_i^{v_t})).
\end{equation}
And then, the actor take the state $s_i^t$ as input and decide the next view proposal $v_{t+1}$ as the action (\emph{i.e.}, $v_{t+1}=\mathcal{A}(s_i^t)$). For the general PPO algorithm with the reward $r_i^t$ for $t$-th step, the advantage estimator $\hat{A}_i^t$ can be expressed as:
\begin{equation}
    \hat{A}_t=-V(s_i^t)+r_i^t+\gamma r_i^{t+1}+\dots+\gamma^{T-t} r_i^T,
\end{equation}
where $V(s_i^t)$ is the learned state-value function, $\gamma \in (0,1)$ is a pre-defined discount factor, $T$ is the maximum length of the input sequence. The principle behind it is straightforward -- the current action should not only benefit the next step but also contribute to the overall goal. However, in this work, aiming at achieving reliable prediction with the least number of steps, we only focus on the profit at the very next step, \emph{i.e.}, we set $\gamma=0$. The advantage estimator we use can be formulated by:
\begin{equation}
    \hat{A}_i^t=-V(s_i^t)+r_i^t.
\end{equation}
After that, we denote the prediction probability of $v_t$ by $\mathcal{A}(v_t|s_i^t)$. Then the clipped surrogate objective is:
\begin{equation}
    L_{CLIP}=\frac{1}{B}\sum_{i=1}^B\sum_{t=2}^T min\left\{\frac{\mathcal{A}(v_t|s_i^t)}{\mathcal{A}_{old}(v_t|s_i^t)}\hat{A}_i^t,clip(\frac{\mathcal{A}(v_t|s_i^t)}{\mathcal{A}_{old}(v_t|s_i^t)},1-\epsilon,1+\epsilon)\hat{A}_i^t\right\},
\end{equation}
where $\mathcal{A}_{old}(\cdot)$ stands for the actor before update, and $\epsilon\in(0,1)$ is a hyper-parameter. Note that $t$ starts from $t=2$ since the first view is randomly selected. Finally, the overall objective of \textbf{Stage II} can be expressed as:
\begin{equation}
    L_{Stage2}=L_{CLIP}-c_1 L_{VF}+c_2 L_E,
\end{equation}
where $L_{VF}=\frac{1}{B}\sum_{i=1}^B\sum_{t=2}^T (V(s_i^t)-V^{target}(s_i^t))^2$ is the squared-error loss suggested by~\cite{schulman2016high}, and $L_E=\frac{1}{B}\sum_{i=1}^B\sum_{t=2}^T S_{\mathcal{A}}(s_i^t)$ is the entropy bonus following~\cite{williams1992simple,mnih2016asynchronous}. $c_1$ and $c_2$ is hyper-parameters to balance the three loss components.

\subsection{Training and Inference Details}

\textbf{Stage I.} Similar to the training of most FGVC models, the backbones (ResNet~\cite{he2016deep}, DenseNet~\cite{huang2017densely}, EfficientNet~\cite{tan2019efficientnet}, RegNet-Y~\cite{radosavovic2020designing}, and ViT~\cite{dosovitskiy2020image}) are all first initialized with ImageNet pre-trained weights. We use SGD optimizor with a momentum of $0.9$ and the cosine learning rate schedule~\cite{loshchilov2016sgdr} for optimization. The start learning rate is set to be $0.005$ for the backbone and $0.05$ for the other components. The input images are random-resize-cropped to $224\times224$. The model is trained for $60$ epochs. The temperature $h$ for entropy maximization is set to be $2$.

\textbf{Stage II.} We use the Adam optimizor with $\beta_1=0.9$, $\beta_2=0.999$, and the cosine learning rate schedule~\cite{loshchilov2016sgdr} for optimization. The start learning rate is set to be $0.00005$ for the backbone and $0.0005$ for the other components. The input images are random-resize-cropped to $224\times224$. The model is trained for $15$ epochs. The hyper-parameters $\epsilon$, $c_1$, and $c_2$ are set to be $0.2$, $0.5$, and $0.01$, respectively.

\textbf{Stage III.} Similar to $Stage I$, the SGD optimizer with a momentum of $0.9$ and the cosine learning rate schedule~\cite{loshchilov2016sgdr} is adopted for optimization. The start learning rate is set to be $0.005$ for both the backbone and other components. The input images are random-resize-cropped to $224\times224$. The model is trained for $60$ epochs.

\textbf{Inference.} The input images are first resized to a fixed size $256\times256$ and then center-cropped to $224\times224$.

\section{Additional Experimental Results}

\subsection{Aggregator Architecture}

Here we conduct ablation studies to select the best aggregator architecture. There are four options being evaluated: multiple fully connected layers, LSTM~\cite{hochreiter1997long}, GRU~\cite{chung2014empirical}, and self-attention~\cite{vaswani2017attention}. Specifically, we train $T$ fully connected layers for each step with different channel numbers for the multiple fully connected layer scheme. The feature sequence $\{x_i^1,\dots,x_i^T,\dots,x_i^T\}$ is concatenated and processed by the corresponding fully connected layer. As for the self-attention architecture, we adopt $4$ multi-head attention layers with $8$ attention heads. We experiment with only \textbf{Stage I} which is enough to reveal the option with the best feature aggregation ability. The experimental results in Table~\ref{tbl:architecture} suggest that GRU can deliver the best performance.

\begin{table*}[htbp]
\caption{Ablation studies about different aggregator architectures. The best results are marked in \textbf{bold}.}
\vspace{0.3cm}
\label{tbl:architecture}
\begin{adjustbox}{width=0.4\linewidth,center}
\begin{tabular}{c c}
\toprule[1pt]
\textbf{Architecture} & mAcc. ($\%$) \\
\midrule[1pt]
Multiple FC Layer & $81.97\pm0.8$\\
LSTM & $83.76\pm0.5$\\
GRU & $\mathbf{84.18\pm1.1}$\\
Self-Attention & $82.82\pm1.8$\\
\midrule[1pt]
\end{tabular}
\end{adjustbox}
\end{table*}

\subsection{Learning Rate}

Here we carry out ablation studies about learning rates at each training stage. The experiments are conducted in a stage-by-stage manner, \emph{i.e.}, the optimal learning rate is selected for each stage according to the model performance at the current stage, and once we finish the current stage, we will move to the next stage with the best model at the current stage as initialization. The experimental results are reported in Table~\ref{tbl:stage1_lr},~\ref{tbl:stage2_lr}, and~\ref{tbl:stage3_lr} for three stages respectively. Note that we only use mAcc for evaluation in \textbf{Stage I} since there is no active view selection yet. Finally, the optimal learning rates for the three stages are $0.05$, $0.0005$, and $0.005$, respectively.

\begin{table*}[htbp]
\caption{Ablation studies about the learning rate of \textbf{Stage I}. The best results are marked in \textbf{bold}.}
\vspace{0.3cm}
\label{tbl:stage1_lr}
\begin{adjustbox}{width=0.4\linewidth,center}
\begin{tabular}{c c}
\toprule[1pt]
\textbf{Learning Rate} & mAcc. ($\%$) \\
\midrule[1pt]
$0.1$ & $19.45\pm8.7$\\
$0.05$ & $\mathbf{84.18\pm1.1}$\\
$0.02$ & $80.81\pm0.5$\\
$0.01$ & $68.16\pm0.3$\\
$0.005$ & $37.54\pm0.6$\\
\midrule[1pt]
\end{tabular}
\end{adjustbox}
\end{table*}

\begin{table*}[htbp]
\caption{Ablation studies about the learning rate of \textbf{Stage II}. The best results are marked in \textbf{bold}.}
\vspace{0.3cm}
\label{tbl:stage2_lr}
\begin{adjustbox}{width=0.8\linewidth,center}
\begin{tabular}{c c c c}
\toprule[1pt]
\textbf{Learning Rate} & mAcc. ($\%$) & w-mAcc. ($\%$) & Step$2$-Acc. ($\%$)\\
\midrule[1pt]
$0.001$ & $84.71\pm1.1$ & $86.10\pm1.0$ & $85.19\pm0.9$\\
$0.0005$ & $\mathbf{85.31\pm1.6}$ & $\mathbf{86.57\pm1.6}$ & $\mathbf{85.63\pm1.7}$\\
$0.0002$ & $84.66\pm1.0$ & $85.79\pm1.0$ & $84.69\pm1.0$\\
$0.0001$ & $84.49\pm1.2$ & $85.29\pm1.5$ & $84.16\pm1.5$\\
$0.00005$ & $84.53\pm1.2$ & $85.22\pm1.6$ & $83.85\pm1.6$\\
\midrule[1pt]
\end{tabular}
\end{adjustbox}
\end{table*}

\begin{table*}[htbp]
\caption{Ablation studies about the learning rate of \textbf{Stage III}. The best results are marked in \textbf{bold}.}
\vspace{0.3cm}
\label{tbl:stage3_lr}
\begin{adjustbox}{width=0.8\linewidth,center}
\begin{tabular}{c c c c}
\toprule[1pt]
\textbf{Learning Rate} & mAcc. ($\%$) & w-mAcc. ($\%$) & Step$2$-Acc. ($\%$)\\
\midrule[1pt]
$0.01$ & $85.49\pm0.5$ & $87.12\pm0.7$ & $86.24\pm0.8$\\
$0.005$ & $\mathbf{86.07\pm0.5}$ & $\mathbf{87.66\pm0.5}$ & $\mathbf{87.20\pm0.4}$\\
$0.002$ & $85.51\pm0.7$ & $86.69\pm0.9$ & $85.70\pm0.8$\\
$0.001$ & $85.34\pm0.8$ & $86.54\pm0.9$ & $85.47\pm0.6$\\
$0.0005$ & $85.20\pm0.7$ & $86.48\pm0.7$ & $85.53\pm0.4$\\
\midrule[1pt]
\end{tabular}
\end{adjustbox}
\end{table*}

\subsection{Temperature for Entropy Maximization}

Here we discuss the effect of the temperature $h$ for entropy maximization. Instead of directly maximizing the entropy of model prediction, we apply a temperature $h>1$ to smooth the prediction distribution as the optimization target. In this way, we are able to explicitly control the degree of entropy maximization constraint. Note that $h=1$ is equivalent to the entropy maximization being disabled. In addition to applying a consistent $h$, we also experiment with a series of exponentially decreased $h$ starting from $5$ -- $[5,3,2,1.5,1.25,1.125,1.0625]$, which follows our intuitive conjecture that the model should yield more confident predictions with more visual inputs. Finally, according to Table~\ref{tbl:temperature}, we choose $h=2$ since it leads to two of the three best results.

\begin{table*}[htbp]
\caption{Ablation studies about different temperature $h$ for the entropy maximization constraint. The best results are marked in \textbf{bold}.}
\vspace{0.3cm}
\label{tbl:temperature}
\begin{adjustbox}{width=1\linewidth,center}
\begin{tabular}{c c c c}
\toprule[1pt]
\textbf{Temperature} $h$ & mAcc. ($\%$) & w-mAcc. ($\%$) & Step$2$-Acc. ($\%$)\\
\midrule[1pt]
$1$ & $85.91\pm0.9$ & $87.07\pm0.9$ & $86.07\pm0.9$\\
$2$ & $86.07\pm0.5$ & $\mathbf{87.66\pm0.5}$ & $\mathbf{87.20\pm0.4}$\\
$5$ & $\mathbf{86.18\pm0.1}$ & $87.30\pm0.3$ & $86.36\pm0.2$\\
$10$ & $85.54\pm0.1$ & $86.64\pm0.7$ & $85.70\pm0.8$\\
$20$ & $85.52\pm0.5$ & $86.77\pm0.7$ & $86.05\pm0.7$\\
$[5.0, 3.0, 2.0, 1.5, 1.25, 1.125, 1.0625]$ & $85.92\pm0.6$ & $87.24\pm0.6$ & $86.54\pm0.6$\\
\midrule[1pt]
\end{tabular}
\end{adjustbox}
\end{table*}

\subsection{Training Scheme}

In this paper, we adopt a multi-stage training scheme for best performance. However, an end-to-end training strategy is also practicable for the proposed framework. Therefore, a comparison of these two schemes is carried out. Specifically, we merge the three training stages into one, \emph{i.e.}, the model is optimized via $L_{CE}$ for recognition, $L_{EM}$ for smooth prediction, and $L_{PG}$ for next-view selection together at each iteration. The model is trained for $60$ epochs. The experimental results are reported in Table~\ref{tbl:training_scheme}. The stage-by-stage scheme outperforms the end-to-end scheme with significant margins, which indicates the necessity of adopting three training stages separately for their different objectives.

\begin{table*}[htbp]
\caption{Ablation studies about different training schemes. The best results are marked in \textbf{bold}.}
\vspace{0.3cm}
\label{tbl:training_scheme}
\begin{adjustbox}{width=0.8\linewidth,center}
\begin{tabular}{c c c c}
\toprule[1pt]
\textbf{Training Scheme} & mAcc. ($\%$) & w-mAcc. ($\%$) & Step$2$-Acc. ($\%$)\\
\midrule[1pt]
End-to-End & $81.41\pm0.7$ & $85.81\pm0.7$ & $85.73\pm0.9$\\
Stage-by-Stage & $\mathbf{86.07\pm0.5}$ & $\mathbf{87.66\pm0.5}$ & $\mathbf{87.20\pm0.4}$\\
\midrule[1pt]
\end{tabular}
\end{adjustbox}
\end{table*}

\section{Further Discussion}

\subsection{Limitation}

The limitations of this work are mainly two-fold. Firstly, for academic purposes only, the collected MvCars is relatively small-scale under the current trend of developing large-scale datasets, making it insufficient to support mature commercial applications. Here we only try to break the ice, hoping to arouse the attention of the FGVC community so as to emerge more and deeper research achievements beyond the $2$D scenario. Secondly, the proposed method adopts GRU~\cite{chung2014empirical} for feature aggregation, which makes it order-sensitive -- different input orders of the same contents may change the prediction results. A significant further impact is that the model performance still lower than the upper bound with a margin of $\sim4.\%$ at the last step (\emph{i.e.}, all visual information is acquired). However, an ideal recognition model based on sequence inputs should be order-invariant rather than forgetting early inputs. Therefore, developing better feature aggregation techniques may be a meaningful future direction.

\subsection{Broader Impact}

Fine-grained visual classification has demonstrated its application value in many fields, \emph{e.g.}, intelligent retail, intelligent transportation, automatic biodiversity monitoring, and many more~\cite{wei2021survey}. Recently, with the development of hardware equipment, portal devices and embodied AI agents tend to be the carrier of computer vision algorithms, which put forward requirements to the vision algorithms for the dynamic information processing ability in $3$D environments. However, the advanced FGVC techniques are still limited to processing $2$D static images despite the great success. In this work, with a newly collected testbed and a viable approach, we may motivate other researchers to develop more effective/efficient algorithms or contribute more challenging datasets to the problem raised. Embracing the coming approaching trend, we believe this could be a new stage for fine-grained recognition research and potentially boost other related tasks, \emph{e.g.}, active fine-grained retrieval, fine-grained $3$D object generation, \emph{etc.}

On the other hand, as the common negative impact for all FGVC tasks, it may be used for military purposes or facilitate criminal behaviours. Besides, the proposed method also suffers the risk of potential adversarial attacks due to the inherent characteristics of deep-neural-network-based models. However, we believe the consequent benefits outweigh the potential negative effects.

\end{document}